# Theory of Cognitive Relativity: A Promising Paradigm for True AI

Yujian Li

College of Computer Science, Faculty of Information Technology, Beijing University of Technology.

Email: liyujian@bjut.edu.cn

**Abstract**

The rise of deep learning has brought artificial intelligence (AI) to the forefront. The ultimate goal of AI is to realize machines with human mind and consciousness, but existing achievements mainly simulate intelligent behavior on computer platforms. These achievements all belong to weak AI rather than strong AI. How to achieve strong AI is not known yet in the field of intelligence science. Currently, this field is calling for a new paradigm, especially Theory of Cognitive Relativity (TCR). The TCR aims to summarize a simple and elegant set of first principles about the nature of intelligence, at least including the Principle of World's Relativity and the Principle of Symbol's Relativity. The Principle of World's Relativity states that the subjective world an intelligent agent can observe is strongly constrained by the way it perceives the objective world. The Principle of Symbol's Relativity states that an intelligent agent can use any physical symbol system to express what it observes in its subjective world. The two principles are derived from scientific facts and life experience. Thought experiments show that they are important to understand high-level intelligence and necessary to establish a scientific theory of mind and consciousness. Rather than brain-like intelligence, the TCR indeed advocates a promising change in direction to realize true AI, i.e. artificial general intelligence or artificial consciousness, particularly different from humans' and animals'. Furthermore, a TCR creed has been presented and extended to reveal the secrets of consciousness and to guide realization of conscious machines. In the sense that true AI could be diversely implemented in a brain-different way, the TCR would probably drive an intelligence revolution in combination with some additional first principles.

# 1. Introduction

Alan Turing is widely considered to be the father of theoretical computer science and artificial intelligence (AI) (Beavers2013). By a seminal paper (Turing 1950), he introduced the Turing test to help answer the question "can a machine think?" and started interest to realize a machine with human intelligence. Recently, deep learning, esp. with the success of AlphaGo, has renewed the interest again (LeCun et al. 2015; Mnih et al. 2015; Silver et al. 2016), but still far from the ultimate goal to achieve human mind and consciousness (Lake et al. 2017). In computer science, intelligence that machines display is called AI, or machine intelligence, in contrast to the natural intelligence of humans and other animals. AI research is defined as the study of intelligent agents: any device that perceives its environment and takes actions that maximize its chance of successfully achieving its goals.

Synthetically, the goals of AI research include reasoning, knowledge representation, planning, learning, natural language processing, perception and the ability to move and manipulate objects (Nilsson 1998). The basic claim of the AI field is that human intelligence can be so precisely described that a machine can be made to simulate it. General intelligence is among the field's long-term goals (Kurzweil 2005), drawing upon computer science, mathematics, psychology, linguistics, philosophy and many others. There have been three major approaches to AI: symbolism (Simon 1995), behaviorism (Brooks 1990; 1991a; 1991b), and connectionism (Rumelhart et al. 1986). Symbolism is predicated on the preeminence of reasoning-like process and conceptualization. It aims to build disembodied intelligence from high-level symbolic representations, which can only interact with the world via keyboard, screen, or printer. Moreover, it claims that human intelligence can be realized by a computer program and even that the running of the right algorithms on a computer would give rise to consciousness. Behaviorism denies reasoning and conceptualization (Kirsh 1991), attempting to build embodied (or robotic) intelligence that can interact with the real world instead of the constructed worlds by symbols. Connectionism denies that reason-like processes are preeminent in cognition and that core AI is the study of the concepts underpinning domain understanding (Kirsh 1991). It tries to recognize real patterns and represent mental phenomena using artificial neural networks.

A fundamental problem in AI is that nobody really knows what intelligence is, despite innumerable tests available for measuring it (Gregory 1998). For people, intelligence is a very general mental capability that, among other things, involves the ability to reason, plan, solve problems, think abstractly, comprehend complex ideas, learn quickly and learn from experience (Gottfredson 1997). Although the details are debated about the definition of intelligence, a fair degree of consensus has been scientifically reached from many respects of natural intelligence and machine intelligence. Most generally, intelligence measures an agent's ability to achieve goals in a wide range of environments (Legg & Hutter 2007). In history, there were a lot of philosophical explorations about the nature of intelligence (Sternberg 1990). For example, Plato likened people's intelligence to blocks of wax, differing in size, hardness, moistness, and purity. Thomas Aquinas thought that people's intelligence could not approach the omniscience of God. Immanuel Kant believed that people's intelligence is different in kinds and facets. It should be mentioned that, Newell and Simon (1976) formulated a well-known hypothesis for intelligence in their Turing Award paper. The hypothesis, i.e. the physical symbol system hypothesis (PSSH), states that a physical symbol system has the necessary and sufficient means for general intelligent action. Note that a physical symbol system, also called a formal system, can be a digital computer, which takes physical patterns (symbols), combining them into structures (expressions) and manipulating them to produce new expressions.

The PSSH has been heavily attacked by "Chinese room" (Searle 1980), "nouvelle AI" (Brooks 1990; 1991a; 1991b), and other arguments. These arguments cluster around four main themes (Nilsson 2007), advocating that biological intelligence involves meaningful grounding, non-symbolic processing, brain-style mechanisms, and mindless chemical activity. In particular, the Chinese room thought experiment shows that a physical symbol system is not the sufficient condition for general intelligent action (e.g. understanding, intentionality, mind and consciousness), and the nouvelle AI approach shows that a physical symbol system is not the necessary condition for general intelligent action (e.g. insect-level robotic intelligence without symbolic representations). Thus, a physical symbol system should certainly differ from a biological intelligent system, even though it beats human Go champions (Silver et al. 2016). In reality, existing AI achievements mainly simulate intelligent behavior on computer platforms. They belong to weak AI rather than strong AI. Originally, strong AI is a position that the appropriately programmed computer really is a mind, in the sense that computers given the right programs can be literally said to understand and have other cognitive states (Searle 1980). Another version of strong AI states that the appropriately programmed computer with the right inputs and outputs would thereby have a mind in exactly the same sense human beings have minds (Searle 1992). In contrast, weak AI claims that a computer would not necessarily have a mind and consciousness, even it is a super-intelligent machine like AlphaGo Zero (Silver et al. 2017; 2018). Additionally, strong AI may refer to a machine with consciousness, sentience and mind. And it may also mean artificial general intelligence (or full AI), i.e. a machine with the ability to apply intelligence to any problem, rather than just one specific problem. However, weak AI is known as narrow AI (or applied AI), i.e. AI that focuses only on a limited task.

To make the PSSH stand further, the attacks to it are also refuted (Nilsson 2007). One refutation is focused on what symbols are. In this refutation, symbols are a set of entities that can be physical patterns (e.g., chalk marks). , they can even occur as components of symbol structures. Symbol structures can, and commonly do, serve as internal representations (e.g., mental images) of the environment (Simon 2000). So, the "symbols" are physical objects that represent things in the world, symbols (e.g., "dog") that have a recognizable meaning or denotation, and even more complex symbols that are composed of simple symbols. Thus, a physical symbol system exists in a world of objects wider than just these symbolic expressions themselves, and produces an evolving collection of symbol structures grounded in the objects in the environment through its perceiving and effecting capabilities whenever necessary.

However, is there no difference between a physical object and a physical symbol? In fact, the start of human intelligence is not with symbols. Human neonates cannot speak any word at all. By crying instinctively, infants can express a variety of feelings (Chicot 2015), such as hunger, discomfort, or loneliness. Moreover, they may pay more attention to danger (Erlich et al. 2013), and receive more benefits from positive touch (Field 2002). In Dunstan's theory, infants make sound reflexes between 0-3 months, but more elaborate babbling after 3 months. Hence, in the development of intelligence, humans must have conscious experience before language acquisition. Additionally, in genetic epistemology, Piaget (1972) distinguished among three types of knowledge: physical, logical-mathematical, and social knowledge. Physical knowledge refers to knowledge related to objects in the world, which can be gained through perceptual properties. Logical-mathematical knowledge is abstract and must be invented, but through actions on objects that are fundamentally different from those actions enabling physical knowledge. Social knowledge is culture-specific and can be learned only from other people within one's cultural group.

Since humans acquire physical knowledge prior to the other two types of knowledge, physical objects must be different from physical symbols. Therefore, it still remains to answer many questions concerning the PSSH and AI, such as: 1) What is the difference between physical objects and physical symbols? 2) Can machines have conscious experience about physical objects without physical symbols? 3) What physical forms of language can machines use to develop their conscious intelligence? These questions are significant in the field of intelligence science, which is currently calling for a new paradigm, especially "Theory of Cognitive Relativity (TCR)" (Li 2005).

The TCR aims to elucidate the nature of intelligence with a simple and elegant set of first principles at the system level. These first principles must be fundamental and compatible in all phenomena of intelligence, and cannot be deduced from any other principles in physics, chemistry and biology. Although the set of first principles would not result in anything like Maxwell's equations or mas-energy relation $E = mc^2$, it should capture the nature of intelligence comprehensively in some perspectives between science and philosophy. Moreover, it should make a guide to realization of machines with conscious intelligence, particularly different from humans' and animals'. Only "human-different" AI can be regarded as a genuine innovation, whereas "human-like" AI is just a kind of imitation. Rather than brain-like intelligence (Sendhoff 2009), the TCR indeed advocate a promising change in direction to realize true AI, i.e. artificial general intelligence or artificial consciousness. This realization should be brain-different, not based on imitation of humans and animals. Note the quest for artificial flight succeeded when the Wright brothers and others stopped imitating birds and started using wind tunnels and learning about aerodynamics (Russell & Norvig, 2011).

So far, the TCR has included two first principles: the Principle of World's Relativity and the Principle of Symbol's Relativity. In Section 2, the Principle of World's Relativity is presented about conscious experience, stating that the subjective world an intelligent agent can observe is strongly constrained by the way it perceives the objective world. In Section 3, the Principle of Symbol's Relativity is presented about information expression, stating that an intelligent agent can use any physical symbol system to express what it observes in its subjective world. In Section 4, thought experiments are designed to demonstrate that the two principles are important to understand high-level intelligence and necessary to establish a scientific theory of mind and consciousness. Section 5 is an extraction of insights for true AI. Section 6 is a TCT creed for consciousness studies. Conclusions are drawn in Section 7, anticipating that the TCR would combine some additional first principles to form a new paradigm for intelligence science, and that it would probably drive an intelligence revolution in the future.

## 2. The Principle of World's Relativity

The most important aspect of mind is consciousness and our conscious experience of self and world. Intuitively, there exists an external world around us. The problem of why it looks like our observation is the central issue of this section. First, the human world is defined as the observable part of the universe, with the anthropic principle discussed. Second, an animal world is defined likewise on the basis of scientific facts. Third, the Principle of World's Relativity is presented as a generalization of the the anthropic principle. Last, the principle is applied to make the difference between phyical objects and physical symbols.

### 2.1. Human world

What exactly is the external world around us? A simple anwer is the universe, which may refer to such concepts as the cosmos, the world, and nature (Copan et al. 2004; Bolonkin 2011). The universe is defined as all of space and time and their contents (Zeilik & Gregory 1998), including all forms of matter and energy, and the laws that influence them. Additionally, it often means "the totality of existence", or everything that exists, everything that has existed, and everything that will exist (Schreuder 2014). Nonetheless, it remains to answer the question: why does the universe look like what we observe? A philosophical explanation is the anthropic principle:

***Observations of the universe must be compatible with the conscious and sapient life that observes it.***

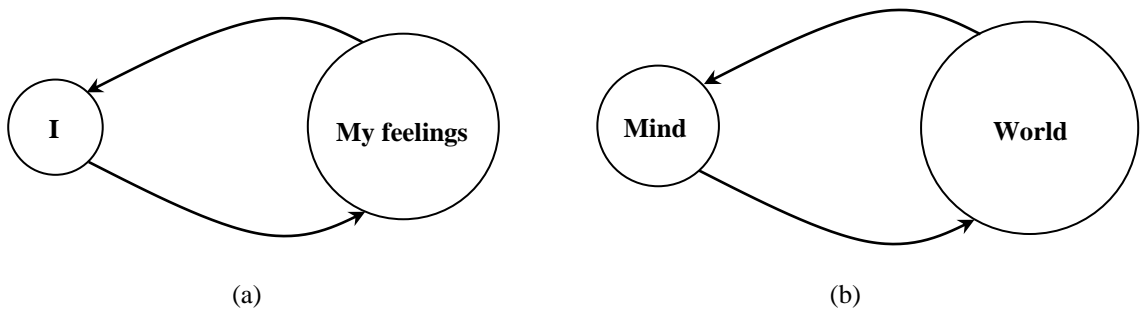

Figure 1. Two strange loops: (a) I-my feelings, (b) mind-world.

The anthropic principle was first articulated by Carter (1974), in order to address the anthropic selection of privileged spacetime locations in the universe and the values of the fundamental constants of physics. The principle has some similar arguments and many variants (Schopenhauer 2016; Barrow 1997). In Carter's view, a weak variant states that our location in the universe is necessarily privileged to the extent of being compatible with our existence as observers, and a strong variant states that the universe (and hence the fundamental parameters on which it depends) must be such as to admit the creation of observers within it at some stage.

It seems that the strong anthropic principle praraphrases Descartes, *cogito ergo mundus talis est*. This strong principle explains why the universe looks like our observation with a strange loop. What is a strange loop? It is a hierarchy of levels, each of which is linked to at least one other by some type of relationship (Barrow 1988; Penrose 1989; Hofstadter 1999; Hofstadter 2013). It may involve self-reference and paradox. Perhaps the best-known strangle loop is the "chicken or the egg" paradox. In processing self-conciousness, the mind perceives itself as the cause of certain feelings, leading to another strange loop that "I" am the source of my feelings (see Figure 1a). Moreover, using the idea of strange loop, we may have another version of the anthropic principle,

***The universe that we find ourselves in is strongly constrained by the requirement that sentient beings like ourselves must actually be present to observe it.***

Acorrdingly, the universe has something to do with our observation. The part that can be observed by humans is called the human world. Note that human observations require the mind to interact with the universe through sensors and effectors. The interactions may affect the way to perceive the external world, and even change it. This can result in a mind-world strange loop (see Figure 1b).

**2.2. Animal worlds**

Perhaps the universe is the same in the mind of each human individual. However, there are so many animal species on the earth. Their observable parts should be different from the human world. The part that can be observed by an animal is called its world. A lot of scientific facts have shown that this anmimal world may not be the same as the human world.

For instance, honeybees are able to compensate for the sun's movement in their navigation. The compensation depends upon a memory of azimuth relative to the honeybees' goal on the previous trip and an extrapolation of the sun's current rate of azimuth movement (Gould 1980). When the sun is removed (e.g. obscured by a cloud, a landmark, or the horizon), the whole sun-centered system of bees is discarded in favor of a backup system — a separate navigational subroutine which is based on the patterns of polarized light generated in the sky by the scattering of sunlight (Dyer & Gould 1981). Since we humans cannot see polarized light directly, the honeybee's world is different from the human world.

Frogs are not concerned with the stationary parts of the world around them. They will starve to death surrounded by stationary food, but can be easily fooled by a piece of dangled meat or any moving small object, and even leap to capture it if it has the size of an insect or worm and moves like one (Lettvin et al. 1959). Of course, we can easily tell an insect or worm from other moving small objects. Hence, the frog's world is different from the human world.

Bats can make use of supersonic wave to navigate and locate. In addition to providing information about how far away a target is, a bat sonar can relay some remarkable details, for example, it conveys information about the relative velocity of a flying insect and its wing-beat, and the size of various features of the target as well as the azimuth and elevation of the target (Suga 1990). In contrast, we humans cannot hear ultrasoundl. Clearly, the bat's world is different from the human world.

Dogs are good at olfactory tracking with a sharp detection threshold for acetic acid that may be 108 times lower than human's threshold (Thesen et al. 1993). With superior olfaction, they can distinguish between more than two million smells, and even determine a difference in the concentration of scent in the air above two consecutive footprints. Clearly, we humans cannot do this at all. It goes without saying that the dog's world is different from the human world.

In addition, snakes can use infrared-sensitive receptors to "see" the radiated heat of warm-blooded prey. Elephants can use ears to hear infrasonic waves. Sharks can use ampullae of Lorenzini [1] to sense elecric fields. Without doubt, their worlds are also different from the human world.

## 2.3. World's relativity

On the basis of scientific facts, it has been shown that the world an animal can observe generally differ from the human world. What about an intelligent machine? Can it observe a world that differs from humans' and animals'?

In artificial intelligence, an agent may refer to a human, an animal, or a machine. If the objective world denotes the universe-in-itself, the world-in-itself, or the ontological world, then the way an agent perceives it may have an effect on the part that the agent can observe. This part is defined as the subjective world of the agent. Accordingly, an agent's subjective world may be greatly affected by the way it perceives the objective world. Therefore, the anthropic principle can be generalized to the **Principle of World's Relativity**:

***The subjective world an intelligent agent can observe is strongly constrained by the way it perceives the objective world.***

Seemingly, the Principle of World's Relativity requires that an intelligent agent should have a mind to observe its subjective world. But it does not deny mindless intelligence, which may exist in birds, lower animals, robots, and even plants. In ethology, birds often exhibit non-symbolic intelligence by carrying out a fixed action pattern (FAP), i.e., a hard-wired and instinctive behavioral sequence that is indivisible and runs to completion (Campbell 1996). A FAP (e.g., a mating dance or an egg rolling process) can be triggered by a sign stimulus, but it looks like an utterly mindless intelligent behavior. Additionally, both single-celled and multi-celled animals survive and reproduce very well without any nervous system at all, and "lower animals," even insects, organize into thriving societies without any symbols, logic, or language, bee dancing and birdsong notwithstanding (Pollack 2006). Additionally, nouvelle AI aims to use inexact and incomplete knowledge to produce robots with intelligence levels similar to insects (Copeland 2015), it tries to get control actions adaptively and mindless intelligence (e.g. more complex behaviors like chasing a moving object) emerge organically from simple behaviors (e.g. like collision avoidance and moving toward a moving object) through interactions with the real world. Finally, plants can have an mindless ability to sense and respond to the environment to adjust their morphology, physiology, and phenotype accordingly (Trewavas 2005). For instance, the dodder (Trewavas 2002), a parasitic plant, assesses the exploitability of a new host within an hour or two of its initial touch contact. If insufficient, it continues searching for other, more profitable, hosts. Otherwise, it coils about the host with a particular number of coils (and eventually suckers) that depends on the assessed future return, and begins to take its host's resources several days later.

Since the Principle of World's Relativity is fundamentally important to understand an agent's subjective world, it should be selected as a first principle to establish a scientific theory of mind and consciousness. According to it, the subjective world only reflects a part of the objective world, like what a blind man feels about the nature of an elephant. The agent can construct its subjective world from the objective world, with its body bridging the two worlds and dissociating them at the same time. Moreover, the subjective world must be materially distinct from the objective world, because the former exists inside the body, with the later outside. The relationship between the two worlds are illustrated in Figure 2.

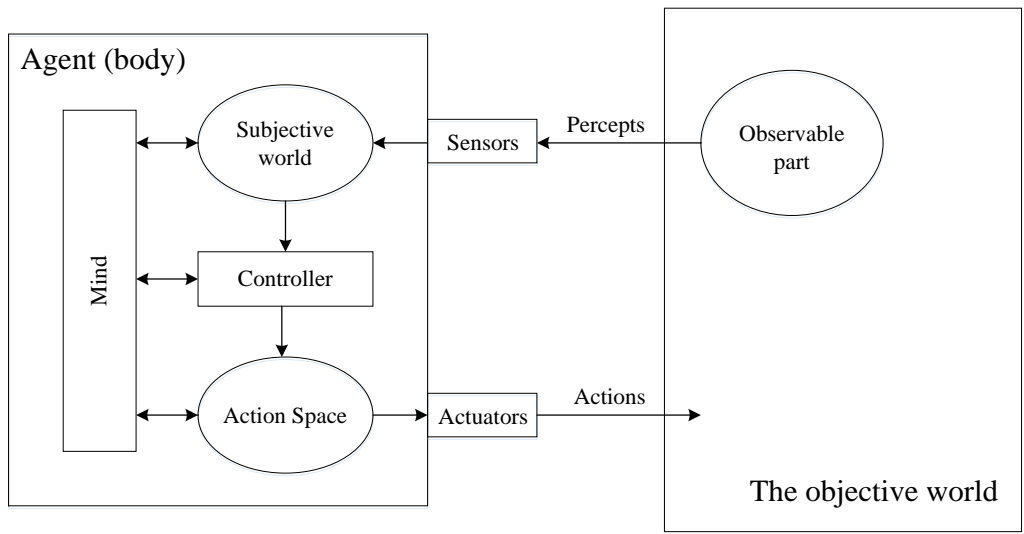

Figure 2. The relationship of an agent's subjective world with the objective world

Because the subjective world is inside an agent's body, it could not be identical to the objective world outside. Otherwise, some conflicts may turn up. For instance, people can observe a subjective house in their brains, but the subjective house may not have the same size as the objective house they are seeing, because the brains are too small to contain it. In a logical view, this means that the subjective world of an agent is something emergent from the objective world through the agent's bodily interactions with it, such as perception and motion. Moreover, the subjective world should be virtual when compared with the reality of the objective world. It can be said that, the subjective world is a virtual reality of the objective world, whereas the reality may be relative to the agent, and vary with its kind.

In Figure 2, the agent seems to have a mind, but this may not be the case. Conceptually, it is an autonomous entity that observes through sensors and acts upon an environment using actuators and directs its activity towards achieving goals. Theoretically, agents care often grouped into five classes: simple reflex agents, model-based reflex agents, goal-based agents, utility-based agents, and learning agents (Russell & Norvig 2011). For example, a thermostat is considered a simple agent. A complex agent may learn or use knowledge to achieve its goals. More generally, an agent is anything that can be viewed as perceiving its environment through sensors and acting upon that environment through actuators. It may be hardware, software, and their combination. In case of a mind, it can use the mind to monitor what happens and to improve its intelligence. Hence, there can be an additional class: minded agents (or conscious agents).

Note that for an intelligent agent, the perception is merely a window to observe the objective world through sensors. Through it, the agent would never be able to get a complete view of the objective world. As a special case, the agent cannot use one sensor to perceive the sensor itself. For example, we cannot use one of our eyes to see itself, unless by instruments.

**2.4. Difference between physical objects and physical symbols**

From the viewpoint of the PSSH, physical objects are not told from physical symbols. The world consists of physical objects and physical symbols. The symbols may include physical objects, simple symbols and complex symbols, even anything that we humans have and use every day of our lives (Newell 1980). Thus, the world an agent can observe is not subjective, which easily confuses the objective world. But according to the Principle of World's Relativity, the world an agent can observe is subjective. The subjective world is greatly dependent on the way it perceives the objective world. If two agents are equipped with different kinds of perceiving sensors, the subjective worlds they can observe may not be the same in general. They may even take an identical physical entity (or stimulus) as different objects. For example, grapes are people's fruit but dogs' poison [2]. Therefore, physical objects are relative to different kinds of agents, and they may compose different subjective worlds. This means that physical object should be different from symbols. However, what exactly is the difference between them?

In practice, a symbol must be implemented in a physical form. In this sense, it is a physical object, and can be called a physical symbol. But in theory it is merely an abstract entity that has no physical properties. Actually, a pure symbol, e.g., "0" and "1", cannot be said hard or soft. Moreover, from a symbol, nobody can get any real experience (or qualia) about the physical object that it refers to. Imagine what an "armadillo" is. Will you get an instance of it? The answer is "no", unless you have a direct perception. The qualia of a real armadillo is beyond any symbolic description of it. By perception, an agent can observe physical objects through sensors to construct its subjective world. The subjective world is composed of the percepts (e.g. mental faces). Usually, these percepts are associated with certain physical objects (e.g. real faces) in the objective world. Nevertheless, this does not imply that the physical objects are always symbols. Only by definition can they be symbols. Without definition, they cannot be taken as symbols. Logically, any physical object can be defined as a symbol to designate some other physical objects, a kind of things, a mental event, even an abstract property, etc. Only as a symbol can it refer to some physical entity (or stimulus) other than itself. In reality, symbols are people-invented things. Essentially, they are a selective collection of physical objects that denotes something else by definition (or convention). Furthermore, they should have relatively simple structures for easy use in representations, computations and communications. Additionally, a same symbol can be implemented in any physical forms, such as audio, visual, tactile, gesture, and even radio.

Finally, it should be noted that, a physical symbol cannot refer to the physical object that implement it. Otherwise, the object is just itself, no longer a symbol. Therefore, a physical symbol system cannot use symbols to recursively represent all physical objects in the objective world, and it is not the sufficient means for general intelligent action.

## 3. The Principle of Symbol's Relativity

Although reflexes play a role in growth, human intelligence largely starts with conscious experience about physical objects through perceiving sensors (e.g. eyes, ears, nose, tongue and body), leading to gradual acquisition of physical knowledge. From the viewpoint of genetic epistemology (Piaget 1972), people can further use physical symbols to develop and improve their conscious intelligence by inventing abstract logical-mathematical knowledge through actions, and by learning culture-specific social knowledge from other people. According to the Principle of World's Relativity, intelligent machines can also have conscious experience about physical objects without physical symbols. But their physical objects may be different from those in the human world. Since the physical symbols are a selective collection of physical objects that denotes something else, the machines will probably be unable to use the same physical forms of human language at all. The central issue of this section is: what physical forms of language they can use to develop and improve their conscious intelligence based on physical knowledge. First, the symbol grounding problem is considered on symbolic meaning. Second, the Principle of Symbol's Relativity is derived from life experience. Third, the principle is applied to analyze limits on symbolic AI. Last, it is combined with the Principle of World's Relativity to analyze limits on computational intelligence and computationalism.

### 3.1. Symbol grounding problem

Generally, the meaning of a symbol is its referent, i.e. the thing that it refers to. A compound of symbols may have an idiomatic referent, which is not always the sum total of the meanings of the symbols. However, a referent could be distinguished from the meaning. For example, Abraham Lincoln and the sixteenth President of the United States both have the same referent, but not the same meaning.

The problem of how symbols get their meanings involves the symbol grounding problem (Harnad 1990). The mediation of the mind would play a critical role in the intentional connection between symbols and any intended referents. The meaning of a symbol on a page is ungrounded. In contrast, the symbols one does understand are grounded in his head. The brain would make them become meaningful thoughts by the mind.

To avoid infinite regression, a symbol grounding process is suggested to require two properties (Harnad 1990; Cangelosi & Harnad 2001): 1) capacity to pick referents, and 2) consciousness. Obviously, a symbol system alone cannot have the body-dependent capacity of picking out referents, except the part of pure computation. To be grounded, it would have to be augmented with non-symbolic sensorimotor capacities to interact with the objective world. In other words, it should be an embodied agent. For this agent, symbolic meaning could be ultimately grounded in its capacity to detect, categorize, identify, and act upon the things that symbols and strings refer to. Since an unconscious robot zombie could be possible to pass the Turing test indistinguishably from us for a lifetime, groundedness should not be a sufficient condition for meaning. Thus, consciousness is required as a secondary property. In the Chinese room argument, even Searle has appealed to consciousness when pointing out that the Chinese symbols would be meaningless to him. Otherwise, meaning could be argued to go on in his head, but simply not being conscious of by himself.

Surely, meaning needs grounding. But grounding is not meaning. It is an input-output performance function related to extenal objects, sensory organs, internal symbols, mental states, and intentional actions. One may use sensory organs to perceive the obejctive world, mental states to represent extenal objects, internal symbols to formulate thoughts, and responding effectors to execute intentional actions.

### 3.2. Symbol's relativity

Commonly, symbols are used to express meanings in a context of groundedness and consciousness. Moreover, a system of symbols can make up a language. Language is a basic tool for thinking and communication in human society. In different countries, people generally speak different languages. There are about 5000~7000 languages spoken all over the world, 90% of them used by less than 100000 people. As estimated by UNESCO (The United Nations' Educational, Scientific and Cultural Organization), the most widely spoken languages are: Mandarin Chinese, English, Spanish, Hindi, Arabic, Bengali, Russian, Portuguese, Japanese, German and French. In practice, a language usually takes forms of speech and text, but it can also be encoded into whistle, sign, braille, or gesture. This leads to an interesting question, can machines think in language of other forms, e.g. radio?

In daily life, people are accustomed to thinking and communication in sound language. To all appearance, a Chinese can think in Chinese, an American can think in English, a Spanish can think in Spanish, and so on. From the viewpoint of life experience, all these spoken forms of language, even including any other forms such as whistle, sign, braille and gesture, should be equivalent for thinking and communication. This self-evident point can generalize to an important principle, termed the Principle of Language's Relativity (Li 2018), the Principle of Symbolic Relativity" (Li 2005) or the Principle of Symbol's Relativity. The principle may be described as follows,

***All admissible forms of language are equivalent for an intelligent system to think about the world.***

In the Principle of Symbol's Relativity, an admissible form means that the system can use it for thinking, i.e. the formulation of thoughts about the world. Also, the principle can be stated in other words,

***All admissible forms of language are equivalent with respect to the formulation of thoughts about the world.***

Note that it is named with inspiration from the principle of relativity in physics [3], namely,

***All admissible frames of reference are equivalent with respect to the formulation of the fundamental laws of physics.***

That is, physic laws are the same in all reference frames - inertial or non-inertial. By analogy, a language can be regarded as a frame of reference to express information (e.g. thoughts and ideas). In this sense, all forms of language, such as speech, text, whistle, sign, braille and gesture, can be readily understood to have equivalence in expression of the same information. Note that different forms of language may not be as easily implemented as each other (e.g., it is easier to write "US" than "United States"). Actually, they are not equivalent in physical implementation. But they are theoretically equivalent to produce the function of information expression. According to the Principle of Symbol's Relativity, a language (or a symbol system) is independent of its physical forms to express information from perception. No matter what forms it is physically implemented, all the forms are equivalent to formulate thoughts, despite a wide range of difficulties. Therefore, the Principle of Symbol's Relativity can also be stated as follows,

*An intelligent agent can use any physical symbol system to express what it observes in its subjective world.*

Obviously, this statement is somewhat like the PSSH. The difference between them lies in that it tells physical symbols from physical objects, whereas the PSSH takes physical objects as physical symbols. In theory, a symbol is an abstract entity, but in practice it must be implemented in some physical form. This implemented symbol is called a physical symbol, which can be generally defined below.

*A physical symbol is a kind of behavior that an intelligent agent can produce, and something that it can make or choose, to define for referring to anything other than itself.*

Based on the above definition, the physical symbol may also be taken as a physical object for the agent itself and others to perceive again and evoke subsequent action. However, in this case the physical object would have a different meaning than its symbolic referent. For example, a signature of name means more than the name itself. Additionally, in communication of ideas, the physical symbol must be recognized by other agents to designate the same thing.

In the Principle of Symbol's Relativity, the subjective world is constructed from an agent's perception of the objective world, where the two worlds can be bridged by the Principle of World's Relativity. Because of independence and compatibility, the Principle of Symbol's Relativity should also be selected as a first principle. Like the Principle of World's Relativity, it is fundamentally important to establish a scientific theory of mind and consciousness (or intelligence more broadly).

According to the Principle of Symbol's Relativity, animals may produce a symbolic intelligence of communication in a different language from people. For example, it is widely accepted that honey bees can use a dance language to encode the distance and direction of the food. After gaining the information from the dance performed by a forager, recruits can fly to find the food (Frisch 1967; Gould 1975a; 1975b). The dance language is an explanation of how foragers recruit other workers, perhaps with some role of odor (Munz 2005). This language is extremely simple when compared with human language. One may argue that the dance is a kind of behavioral intelligence, rather than a symbolic intelligence. However, according to the definition of physical symbols, it can also be conceived of as a symbol in the case expressing information about the food. In this sense, it is certainly a kind of symbolic intelligence.

## 3.3. Limits on symbolic AI

Many aspects of intelligence can be achieved by the manipulation of symbols. This kind of intelligence is called symbolic intelligence. Symbolic AI is the term for the collection of all AI methods using high-level symbolic representations of problems, logic and search. The Turing test is a satisfactory operational definition of symbolic intelligence (Turing 1950): a computer passes the test if a human interrogator, after posing some written questions, cannot tell whether the written responses come from a person or from a computer (Russell & Norvig, 2011). The approach of symbolic AI is also named GOFAI (Good Old-Fashioned Artificial Intelligence) (Haugeland 1985), e.g. expert systems with a network of production rules.

Based on the PSSH, a physical symbol system can produce any intelligent action. However, can it always catch the exact meaning of a symbol that refers to a physical object?

According to the Principle of Symbol's Relativity, the answer will be "no", because a physical symbol system can only manipulate pure symbols and their combinations. It may have a symbolic world built from keyboard inputs, but cannot construct a subjective world from perception of the objective world.

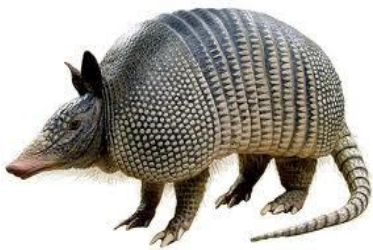

Figure 3. Armadillo

Traditionally, a physical symbol system may be defined as a selective collection of symbols with manipulations on them. In such a system, symbols can be manipulated to produce combinations of symbols, compound symbols, or other new symbols. But elementary symbols have to be defined by people (or other observers) for referring to physical objects. Without observes' explanation, the system can neither relate the elementary symbols to their corresponding physical objects, nor get their exact meaning, for lack of homologous sensors to perceive the objective world. In fact, the subjective experience of physical objects is beyond the expressive capability of symbols. For example, nobody knows what exactly an "armadillo" is, before directly perceiving an instance of it (see Figure 3). As a physical object, the image of an "armadillo" is taken as a symbol by the PSSH. But the image is different from the word "armadillo". A human being can understand the image at once, but not the word. Without seeing the image, even it will be impossible for a human to understand the word "armadillo" exactly, let alone for a physical symbol system. Hence, the PSSH cannot solve the symbol grounding problem substantially. In other words, a physical symbol system cannot get any conscious experience (e.g. qualia) from the objective world, thus it is not the sufficient condition for general intelligent action.

**3.4. Limits on computational intelligence and computationalism**

Human intelligence is non-symbolic in general. Although this non-symbolic intelligence may be non-computational, current AI can be almost considered symbolic or computational. Computational intelligence is the study of the design of intelligent agents (Poole et al. 1998). It aims to address complex real-world problems with a set of nature-inspired computational methodologies and approaches (Siddique, 2013), and tries to produce the learnability of a computer from data or experimental observation, where there might be too complex processes with some uncertainties for mathematical reasoning.

Compared with symbolic AI, computational intelligence is more concerned with the problem of how to produce AI with inexact and incomplete knowledge instead of strict description in formal symbols. It is a useful approach in computational theory of mind (CTM) [8], or computationalism (Scheutz 2003). As a position of strong AI (Searle 1980), the CTM is the philosophical position that human minds are computer programs in essence, or more concretely that the appropriately programmed computer with the right inputs and outputs would thereby have a mind in exactly the same sense human beings have minds (Searle 1999). More generally, the CTM is a family of views holding that the human mind is an information processing system and that cognition and consciousness are a form of computation. It entails the computational theory of cognition (CTC), which provides an explanatory framework of understanding neural networks. According to the CTC, neural activity is computational, and neural computations explain cognition (Piccinini & Bahar 2013), but it leaves open the possibility that phenomenal consciousness could be non-computational (Harnad 1994). However, the CTM asserts that not only cognition, but also phenomenal consciousness (or qualia), are computational. According to the CTM, the mind is not simply analogous to a computer program, but literally a computational system that is physically implemented by neural activity in the brain (Horst 2005). If neural activity is a kind of computation, then the mind, generated by it, should be able to implement in silicon chips, or artificial neural networks. This is a general point of the CTM.

The CTM has been argued against by Searle's Chinese room and Mary's room. Searle (1980) argues that computers cannot be said to have intentionality and understanding, and they are insufficient for the study of the human mind. Mary's room is a thought experiment also known as Mary the super-scientist (Jackson 1982). As a brilliant scientist, Mary knows everything about the science of color perception (e.g. brain states and physical properties), but has never experienced color in a black and white room. The question is: once she experiences color, does she learn anything new? The answer "yes" shows that she can possibly discover some non-physical knowledge only through conscious experience (or qualia). Thus, qualia of seeing color (e.g. red, green and blue) are something nonphysical beyond the interpretability of neural activity. In other words, the computation of neural activity does not explain the human mind wholly.

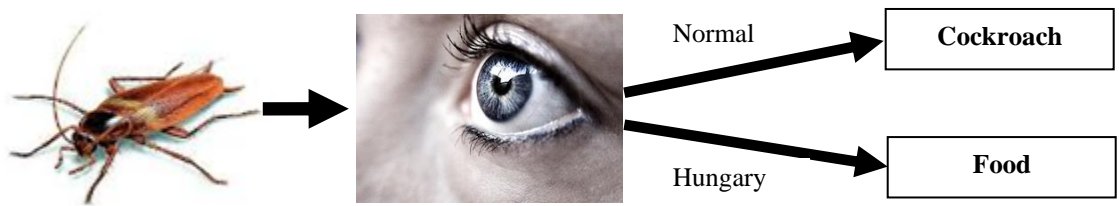

Figure 4. Human perceptions in computational and non-computational views.

According to the Principle of World's Relativity, the human mind can have phenomenal consciousness in non-computational ways. Moreover, the principle claims that phenomena are subjective, and may vary with different kinds of intelligent agents, which can observe the subjective worlds composed of physical objects. The physical objects are strongly constrained by the agent's perception of the objective world. According to the Principle of Symbol's Relativity, a selective collection of them can be defined as symbols to represent any other physical objects. Through computation of these symbols, the agent would be able to behave intelligently in activities of achieving goals. However, physical objects are different from physical symbols. Generally, one cannot catch the exact meaning of a symbol (e.g. "armadillo"). Hence, the super-scientist Mary cannot have a real experience of color based on the science of color perception, unless having chances to seeing color directly. In addition, as a subjective experience of physical objects, qualia may not be computational at all. For example, human perception sometimes requires non-computational processing. That is to say, perception is not always computation. In fact, computation is a transforming process from input to output, theoretically it produces the same output for the same input in general. However, perception is not just a signal-to-symbol transformation. As Figure 4 shows, by seeing people normally recognize a real cockroach as symbol "cockroach" that is uneatable in computational view, but in hungry state they may alternatively take it as "food" in non-computational view. No doubt, hungriness is a desire to eat, involving digestion of food. To all appearances, digestion is not a computational process, but a chemical process that cannot be sufficiently understood in the computational metaphor.

Furthermore, taste of sugar requires non-computational physical/chemical processing that follows the natural laws, beyond computational simulations. From the point of TCR's view, even if computational models could help intelligent agents to construct their subjective worlds from the objective world through physical interactions, the subjective worlds would not be absolutely computational in themselves. Actually, they should always have pertinence to perceiving sensors of the agents and matter distributions in the environments. In addition, the results of computational models have to be explained and understood by humans or some other kinds of agents in their subjective worlds. Therefore, it is impossible to build a unified computational model of perception for all kinds of intelligent agents. This implies that, intelligence, at least the part of conscious experience, may not be a computer program.

## 4. Thought Experiments

The mind-body problem is not only the central issue of philosophy, but also of intelligence science arguably. A final solution to the problem requires a new paradigm, especially the Theory of Cognitive Relativity, which at least includes two first principles: the Principle of World's Relativity and the Principle of Symbol's Relativity. To demonstrate their importance and necessity for solving the mind-body problem, in this section the two first principles are exploited to analyze four thought experiments that focus primarily on the theme of mind, consciousness and intelligence. Respectively, these thought experiments are: 1) brains in a vat, 2) removal of perceptions, 3) robonauts in space exploration; 4) society of artificial mind.

### 4.1. Brains in a vat

In philosophy, the brain in a vat (or brain in a jar) is a well-known thought experiment [4], which raises issues about the mind/world relationship, especially the classical problem of skepticism with respect to the external world in a modern way. This experiment is intended to draw out certain features of human conceptions of knowledge, reality, truth, mind, consciousness and meaning. It outlines a scenario in which a person's brain might be removed from the body and suspended in a vat of life-sustaining liquid. Meanwhile, the nerve endings have been connected by wires to a supercomputer (see Figure 5). The problem is, if the supercomputer were so clever to provide the brain with electrical impulses identical to those it normally receives through the nerve endings, would it get the same view of reality as an embodied brain? In other words, without being related to objects or events in the real world, would the disembodied brain continue to have perfectly normal conscious experiences? For example, the illusion of people, grass, cats, dogs, apples, flowers, trees, houses, the sky, etc.

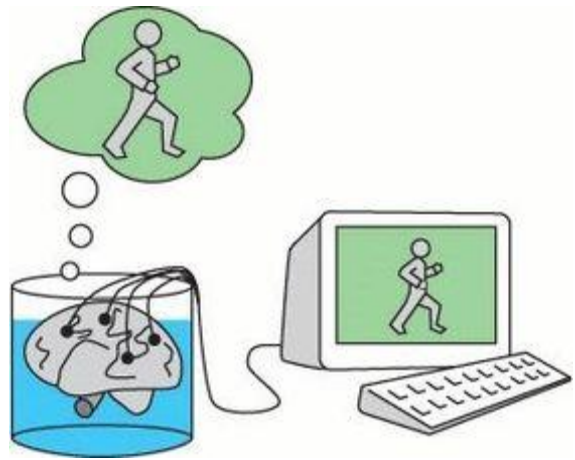

Figure 5. A disembodied brain with the nerve endings connected to a supercomputer [4].

Instead of just one brain, it could be imagined that all human (or sentient) beings are brains in a vat. Moreover, the supercomputer can give them all a collective hallucination, rather than a number of separate unrelated hallucinations. In this case, would the disembodied brains have the consciousness that is perfectly consistent with everything they have experienced?

From the viewpoint of the identical impulses, it seems that the disembodied and embodied brains would have a totally consistent view about their external worlds. However, according to the Principle of World's Relativity, they would not because of different perceiving sensors. Notably, a disembodied brain lacks the connections from the body to it, whereas an embodied brain receives the stimuli from the sensors found in the body. They are neither neuroanatomically nor neurophysiologically similar in their perceptual structures (Heylighen & Apostel 2012). Therefore, the disembodied brain perceives the objective world in a different way from the embodied brain. Accordingly, their subjective worlds would not be the same completely. At least, the disembodied brains cannot think or say anything about where exactly they are. Otherwise, this will lead to some kind of self-refuting arguments, for example, "they are brains in a vat," or "we are brains in a vat" [5]. In addition, the disembodied brains cannot think or say anything about their bodies. Otherwise, this will result in another kind of self-refuting arguments, for example, "they all have bodies," or "we all have bodies". Note that neither can the embodied brains with each in a scull be in a vat, nor can the disembodied brains have any bodies. Therefore, the disembodied and embodied brains must have something inconsistent with their conscious experiences.

**4.2. Removal of perceptions**

In neuroscience, consciousness is hypothetically generated by the interoperation of various parts of the brain, called the neural correlates of consciousness (NCC). The NCC constitute the minimal set of neuronal events and mechanisms sufficient for a specific conscious percept (Koch 2004). Some proponents believe that this NCC interoperation can possibly be emulated by computer systems or cognitive robotics.

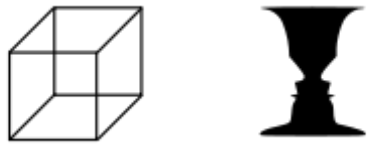

Figure 6. The Necker cube and Rubin vase.

However, according to the Principle of World's Relativity, consciousness is closely related to perception. However, what exactly is their relationship? Clearly, the brain uses perception to understand the environment by interpretation of sensory signals that go through the nervous system. These signals, e.g. light, pressure waves and odor molecules, are related to different types of perception, including vision, hearing, smell, taste, touch, etc. Apart from the passive receipt of the signals, the perception is also shaped by the recipient's learning, memory, expectation, and attention. For instance, the Necker cube and Rubin vase can be ambiguously perceived in more than one way (see Figure 6). This shows that a perceptual system can actively attempt to make sense of its input.

How sensory information influences perception is a vital issue in psychology, science, and philosophy. Even though the information is typically incomplete and rapidly varying, the perceptual systems enable the brain to perceive a stable world around. Obviously, perception is strongly correlated to consciousness. Locke defined consciousness as "the perception of what passes in a man's own mind" [6]. It seems that consciousness must be based on perception. But what is the minimum requirement of perception for emergence of consciousness? The problem leads to an interesting thought experiment, called "removal of perceptions".

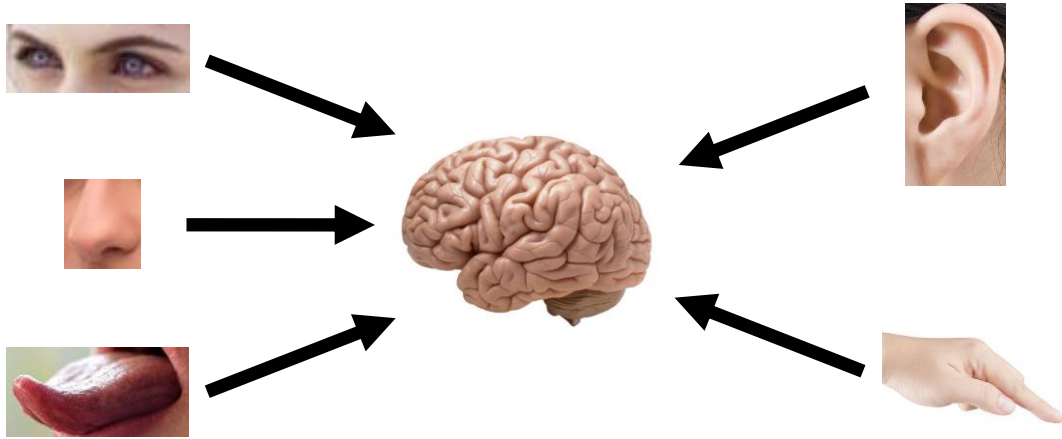

Figure 7. An embodied brain with eyes, ears, nose, tongue, and limbs.

Generally, an embodied brain perceives the environment with sensory organs: eyes, ears, nose, tongue and limbs (see Figure 7). These organs are essential to produce vision, hearing, smell, taste and touch. However, are they all the basic requirements for consciousness? What would happen if some of them were removed? Although all of them play a role in normal consciousness, arguably none of them is indispensable for emergence of consciousness. In reality, a blind person can have consciousness without vision, and a deaf person can have consciousness without hearing. Moreover, a person can also have consciousness after removing other perceptions: smell, taste, touch, etc. Even with all these common perceptions removed, one can still have consciousness in imagination and dream!

What exactly is the relationship between perception and consciousness? According to the Principle of World's Relativity, an intelligent agent should have a subjective world for consciousness. Theoretically, it may observe a multisource subjective world in perceiving the objective world with diverse sensors. Presumably, the simplest subjective world would be at least composed of existence and non-existence about something from one source. Therefore, consciousness needs a minimum requirement of perception to build such a bare-bones subjective world of existence and non-existence. Many types of perception can collectively result in more complex consciousness. A removal of some types will make the consciousness simpler. However, no special perception is indispensable for consciousness. Consciousness can emerge without vision, hearing, smell, taste, or touch. But any type of perception may give rise to consciousness, regardless of what type.

## 4.3. Robonauts in space exploration

With the impressive progress of deep learning (LeCun et al. 2015), esp. the success of AlphaGo (Silver et al. 2016), the interest in building machines that learn and think like people has been excited and renewed again (Lake et al. 2017), despite great challenges to perform a variety of tasks as rapidly and flexibly as people do.

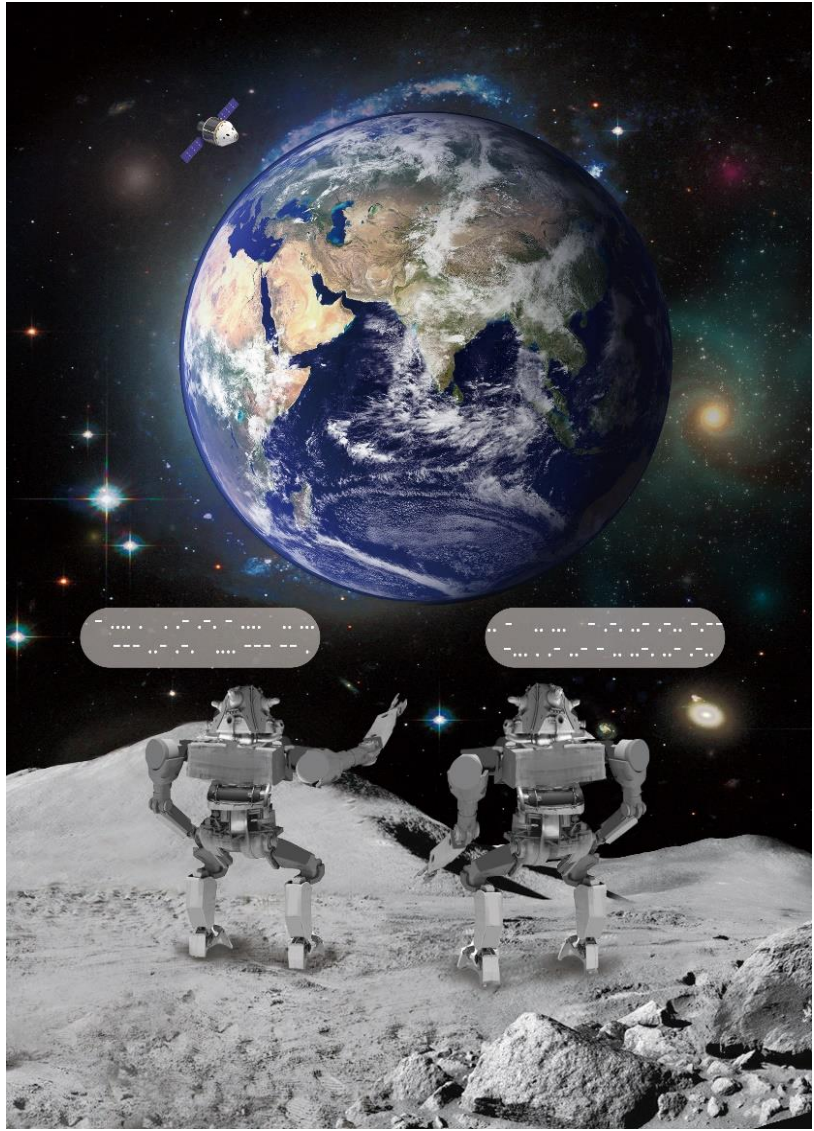

Figure 8. Two robonauts are talking about the earth in a Morse-code radio language on the moon. One says "THE EARTH IS OUR HOME", the other "IT IS TRULY BEAUTIFUL". This is a potential application of the Principle of Symbol's Relativity to artificial intelligence in the future.

What does it mean for a machine to learn and think like a person? Lake et al. (2017) argued that this machine should build causal models of the world, ground learning in intuitive theories of physics and psychology, and harness compositionality and learning-to-learn. They claimed that these key ideas of core ingredients would play an active and important role in producing human-like learning and thought.

Undoubtedly, their claim is attractive for the ultimate dream of implementing machines with human-level general intelligence. However, the claim says little about a person's ability to communicate and think in natural language, which is clearly vital for human intelligence (Mikolov et al. 2018).

It goes without saying that language is an essential ability for human intelligence, e.g. thinking and communication. But is it necessary for intelligent machines to think and communicate in human language? Can they think and communicate in a physical language other than people have ever used? (Li 2018) Moreover, in space exploration of no air, how should a capacity of language be developed for two robot astronauts to talk each other (see Figure 8).

According to the Principle of Symbol's Relativity, language is independent of modality. This would give practical guidance to engineering future generations of intelligent agents. For example, principally robots can think in radio language. These robots would be tremendously useful in space exploration, where for lack of air, radio language is much more convenient for them to talk each rather than sound language. Since no person has an inborn ability to receive and send radio waves, the radio form of language is not admissible for humans. Thus, radio language is a novel and creative idea for robots to think about the world, although radio is certainly very ordinary for information transmission and remote control. Clearly, thinking in radio language (radio thinking), is a people-different way to implement AI. One may argue that, even without language, artificial intelligence could equal or even beat human intelligence in performing such tasks as object recognition (He 2016), video games (Mnih et al. 2015) and board games (Silver et al. 2016). But autonomous robonauts would be more practical on the moon or the other planets if they can use radio language to think, communicate and collaborate. Although Kirobo is the world's first talking robot sent into space [7], it is tasked to be a companion, not an intelligent explorer. With sound language, Kirobo can talk only inside the spacecraft. A solution to this problem would be the use of radio language, by which robonauts could talk outside. Despite that a radio language can be a translation of any human language, in theory it can also be a totally different language with all symbols and words defined arbitrarily and even randomly.

### 4.4. Society of artificial mind

It is well-known that Minsky (1986) published a book titled "The Society of Mind". In his philosophy, a core tenet on the conceptual level is that "minds are what brains do". Moreover, the human mind and any other naturally evolved cognitive systems can be viewed as a vast society of individually simple processes known as agents. Although these processes are not minds, they are fundamental to build minds. Hence, a mind is a society of agents, not the consequence of some basic principle or some simple formal system.

Presently, an agent may refer to a human, an animal, or a machine. Thus, it can be an intelligent machine with artificial mind. A number of minded agents of the same kind would be able to form a society of artificial mind. According to the Theory of Cognitive Relativity, these agents may have a subjective world different from the human world. The question is, what would happen in their society of artificial mind?

The thought experiment "society of artificial mind" can be envisioned in a diversity of cases. For example, in the case that the minded agents could see only ultraviolet, what would happen in their society of artificial mind? In human visual perception, color can be categorized into red, orange, yellow, green, blue, purple, etc. Moreover, these categories are associated with physical objects through the wavelength of the light that is reflected on them. The color of a physical object depends on both the physics of the object in its environment and the characteristics of the perceiving eye and brain. However, in the agents' perception, how could ultraviolet produce color categories in their society of artificial mind? A thoughtful consideration on this question would probably result in a relative science of ultraviolet color to the characteristics of the agent's perceiving eye and "brain", although the Theory of Cognitive Relativity still requires some additional first principles to answer what the agent's "brain" should be.

Another interesting case is that if the minded agents (e.g. aliens) could perceiving nothing in the human world. The question is, with what different kinds of sensors and effectors would they become sufficiently intelligent to establish a scientific theory (e.g. physics) to grasp the nature of the objective world in their society of artificial mind? A deep analysis on this question would probably result in a relative epistemology to the characteristics of the agents' perceiving sensors and responding effectors, which might explain the nature of knowledge relatively, and answer whether knowledge is possible at all.

## 5. Insights for True AI

Recently, AI techniques have experienced a resurgence with advances of deep learning, computer power, and large amount of data. But all AI-related achievements belong to weak AI, far away from strong AI. Although strong AI originally refers to computational theory of mind [8], at present it more likely refers to true AI, i.e. artificial general intelligence (AGI) or artificial consciousness (AC).

In the era after deep learning, it is an urgent desire to realize true AI. The fundamental problem of true AI is whether it can be possible to use abiotic materials to achieve a machine with human mind. Traditionally, even until now, almost all AI researches are preoccupied with simulating the human mind — the zenith of natural intelligence as far as is known. Moreover, these researches are based on the belief that AI systems should understand the objective world in the same way as people do. Thus, discovering the neural basis of mind is arguably regarded as the greatest challenge in cognitive neuroscience (Tenenbaum et al. 2011). The challenge strives to explain the brain's workings with a computational model in theory and simulation (Eliasmith et al. 2012; Gerstner et al. 2012). However, according to the Chinese room argument, the computational model cannot explain the human mind satisfactorily, meaning that it is not a sufficient approach to true AI.

To realize true AI in a different way, in this section the Theory of Cognitive Relativity, or TCR, is exploited to extract insights for artificial general intelligence or artificial consciousness respectively.

## 5.1. Artificial General Intelligence

Artificial general intelligence (Goertzel & Pennachin 2007), is the intelligence of a hypothetical machine that could perform at least the full range of human cognitive abilities. Unlike the breakthrough of weak AI, the AGI makes little progress in the resurgence of AI with deep learning. In early research, many AI pioneers thought that the AGI would possibly exist within just a few decades, to do any work a man can do. Their predictions were the inspiration for character HAL 9000, who was an envisioned embodiment of the AGI. However, the difficulty of making HAL 9000 had been grossly underestimated. It became obvious in the 1970s, increasing criticism of the AGI and pressure to produce useful "applied AI" (i.e. weak AI). In the 1980s, interest in the AGI was revived by Japan's Fifth Generation Computer Project. But the goals of this project were never fulfilled, leading to collapse of confidence in the AGI. By the 1990s, AGI gained a reputation of vain promises and became a topic that would be mentioned reluctantly.

There are a lot of possible reasons for the difficulty of AGI. The first is that computers lack a sufficient scope of memory or processing power. The second is that the level of its relevant complexity may also limit its implementation. The third is the conceptual framework, which should be modified to provide a stronger base for the quest of AGI. Other reasons involve the lack of a complete understanding of human behaviors as well as the need to fully understand the human brain through psychology and neurophysiology.

From the point of TCR's view, the AGI has an unrealistic goal to achieve the human-level AI on a computer platform. According to the TCR, if an intelligent agent perceives the objective world through different kinds of sensors from humans, then it may observe a different subjective world. Since the human subjective world depends strongly on the human brain and body, most of this subjective world is beyond what a computer can do. Although the computer could produce artificial intelligence that are inspired from the human brain, it would not have a human subjective world because of no human body. Therefore, the AGI should not set its goal to achieve the human-level AI on computers. Basically speaking, it requires a revolutionary conceptual framework for intelligent agents that are based on the perceiving sensors, the communicating capabilities, and the executing actuators. On the other hand, probably there will be no such AGI that can perform the best in all respects. In fact, people are not always the most intelligent among biological creatures. Without using tools, people cannot beat owls in catching mice. Moreover, people can neither fly as freely as birds, nor swim as smartly as dolphins. Meanwhile, no matter how to teach owls, birds and dolphins, they will never catch the genuine meanings of differential equations, quantum mechanics and Turing Machines in the way of human understanding.

From the point of the TCR's view, the subjective world is very important for an intelligent to achieve the goals of AGI, but it may not be necessary for a computer to produce weak AI. Without it, a computer program (e.g. AlphaGo) could also exhibit intelligence and even superintelligence in a limited task specific field. However, based on the Chinese room argument (Searle 1980), no matter how intelligently or human-like a program may make a computer behave, it cannot give the computer a "mind", "understanding" or "consciousness", which is a requirement of the AGI. Therefore, in order to have a mind, the AGI should be developed on physical robots that can interact with the objective world through perceiving sensors and responding effectors. Although their subjective worlds could be generally different from the human world, they would also have possibilities to achieve a human-different kind of AGI. A vital issue of TCR would be the problem of how to realize such a kind of AGI without emulating the human brain and body, in particular that it could ultimately establish a scientific theory (e.g. physics) to grasp the nature of the objective world in a different way as people do.

## 5.2. Artificial Consciousness

Artificial consciousness is a hypothetical machine that possesses awareness of external objects, ideas and/or self-awareness. Traditionally, it aims to synthesize human consciousness in an engineered artefact such as computer, but it has not yet been implemented at all in any AI systems. Whether machines may ever be conscious is a controversial question (Dehaene et al. 2017). In order to implement artificial consciousness, there would be many necessary aspects for consideration, such as awareness, memory, learning, anticipation, and subjective experience. However, all these aspects are very difficult to define exactly. Moreover, it is still not easy to draw a clear-cut distinction between conscious and unconscious mental processes. Although the quantity of integrated information is conceived as a theoretic measure that reflects how much consciousness there is (Tonni et al. 2016), mere information-theoretic quantities may not suffice to define consciousness, because it can hypothetically result from specific types of information-processing computations (Dehaene et al. 2017).

Undoubtedly, the conscious mind is the center of human intelligence. But what exactly is consciousness? In general, it is the state or quality of awareness or of being aware of an external object or something within oneself (Gulick 2004). Human consciousness has been defined as: sentience, awareness, subjectivity, qualia, the ability to experience or to feel, wakefulness, having a sense of selfhood, and the executive control system of the mind (Farthing 1992). Animal consciousness can be defined similarly with a lot of evidence (Griffin 2001), it poses the problem of other minds because non-human animals cannot use human language to tell about their subjective experience, such as tastings, seeings, hearings, pains, tickles, itches, and streams of thought. Since subjective experience could be the essence of consciousness, the existence of animal consciousness would never rigorously be known.

From the point of TCR's view, animals may not have the same consciousness as humans' because they generally perceive the objective world in different ways. Accordingly, intelligent machines may also have artificial consciousness that is different from humans' and animals'. Hence, there will be no such universal consciousness that can be defined for all different kinds of intelligent agents. Furthermore, since subjective experience may vary with different types of intelligent agents, a consciousness can only be realized in a special kind of physical systems because some properties of it depend necessarily on physical constitution. Although biological consciousness is physically realized by the hardware of brain and body, this hardware cannot be used to achieve artificial consciousness. Because a computer is not an embodied brain, even if some day it had an artificial consciousness, the consciousness would be different from those that humans and animals have in their brains. According to the TCR, artificial consciousness requires a theory that can be independent of the biological brain, just as artificial flight requires aerodynamics. Note that aerodynamics is a theory independent of the biological bird. That is, it can be described without biological terms. The realization of artificial consciousness should be based particularly on a subjective world built from observations of the objective world. Since the subjective world generally differs from the human world and the animal worlds, this realization is indeed a brain-different approach to true AI, rather than the brain-like intelligence.

From the point of the TCR's view, intelligent machines can have many different levels of artificial consciousness, like biological consciousness. At what level would be related to their ways of perceiving the objective world. The top level might be the AGI. But it is insufficient to realize artificial consciousness merely in computational models. In fact, according to genetic epistemology (Piaget 1972), an agent's conscious intelligence arises from its physical interactions between the subjective world and the objective world. People's logical-mathematical knowledge must be constructed on the basis of physical knowledge. Without physical interactions, even people cannot determine the correctness of their thoughts and computations totally, for example, which geometry is correct, Euclidean or non-Euclidean, in their own reality, although they are so mathematically thoughtful to invent many axiomatic systems for geometries. Therefore, intelligent machines require physical interactions to develop true AI, not only for catching the exact meanings but also for determining the real correctness of their thoughts and computations.

## 6. TCR Creed for Consciousness Studies

The profound and far-reaching implication of TCR lies in that it can help people get rid of the limitations of their own perceptual abilities and go beyond a higher level to understand the relationship between the subjective world and the objective world. Moreover, it may play a practical role of guidance in realization of true AI. Despite that the TCR requires some additional first principles to completion, a TCR creed has been presented (Li 2005) and extended for consciousness studies in area divisions, neuroscientific experiments, theoretical explanations, engineering realizations, etc. The creed mainly includes the following opinions:

1) The study of consciousness should be divided into three directions: human consciousness, animal consciousness, and machine consciousness. Notably, these three directions need to adopt different methods. Neither impose human consciousness readily on animals, nor expect too much that machines will have the same consciousness as humans. Since humans, animals and machines may perceive the objective world in different ways, they can have different subjective worlds in their consciousness, according to the TCR.

2) For human consciousness, the study should focus on the neural mechanisms of how consciousness disorders (e.g. prosopagnosia and schizophrenia) arise and the methods to treat these disorders, because of limitations on normal people. Don't try hard to build a so-called brain model to explain human consciousness in a computational way. Actually, the best explanation of human consciousness is the human brain itself, not its model. To duplicate the performance of the human brain largely, such a model can be extremely complex, and will be hard, or even impossible for us to understand (Churchland & Sejnowski 1988). Even if the model could have a consciousness, perhaps the consciousness would not be the human consciousness at all. In fact, it is more likely to simulate the brain activities of a lower animal, rather than any consciousness.

3) For animal consciousness, the study should focus on how the worlds observed by animals differ from the human world and on how they communicate with each other, and then on how their different kinds of conscious intelligence are produced by the organs, nuclei, and functional areas of the nervous systems. Don't be too enthusiastic in search for the neural counterparts of the outside world or the neural correlates of consciousness (i.e. NCC) for a specific conscious percept (Koch 2004). On the one hand, it could be very hard to confirm the existence of these counterparts or the NCC due to their widely (even wholly) distributed representations. On the other hand, in case of existence they still mean almost nothing to different kinds of individuals, and they may have cerebral positions and firing patterns that vary with different experiences of individuals of the same kind.

4) The most difficult problem of consciousness is to explain how physical processes in the brain give rise to subjective experience. It is also called the hard problem (Chalmers 1995a), which constitutes a real conundrum of mind. This problem seems to defy the possibility of a scientific explanation (Chalmers 1995b). According to the Principle of World's Relativity, subjective experience depends greatly on perception. The hard problem may vary with different kinds of conscious agents. Therefore, it cannot be completely solved by the brain mechanisms. At least, the hard problem of conscious machines needs to be answered in a different way they perceive the objective world.

5) The most important thing for solving the problem of consciousness is to create intelligent machines with artificial consciousness, instead to discover the neural mechanisms of biological consciousness. Although we are still not clear the neural mechanisms of how birds fly, the fact that we can make airplanes means we have already mastered the secrets of flight. Likewise, when we can make conscious machines, it will also mean that we have revealed the secrets of consciousness. Therefore, it can be anticipated that conscious machines will be created before the neural mechanisms of consciousness are revealed, just like the creation of flying machines prior to the neural mechanisms of biological flights.

6) In the development of true AI, it is not a choice of must to employ the brain-like approach (Sendhoff et al. 2009). According to the TCR, the subjective world of a non-human animal is generally different from humans', and that of a machine can be more different from humans'. Hence, true AI could be realized in a brain-different way. This implies that the brain-like intelligence would not be necessary to achieve the goal of true AI, particularly in the present age with insurmountable technological obstacles. For example, it would be a great obstacle to realize a brain-like state of pain by emulating some distinctive kind of C-fiber neural activity, even based on functionalism of the mind [9], which permits multiple realizability of mental states (beliefs, desires, being in pain, etc.).

7) There may be many approaches to realization of conscious machines, but the technological levels of perceiving sensors and pattern recognition largely determine the limits of their subjective worlds (i.e. what these machines can observe), and the limits of their conscious intelligence. According to the Principle of World's Relativity, the subjective world an intelligent machine can observe is strongly constrained by the way it perceives the objective world. Thus, the machine can be equipped with people-different sensors, in order to be able to see polarized light and electromagnetic field, to hear ultrasound and infrasound, to detect ultraviolet and infrared, and so on. According to the Principle of Symbol's Relativity, an intelligent machine can use any physical symbol system to express what it observes in its subjective world. Thus, the machine can be equipped with people-different language (e.g. radio language), in order to be able to formulate thoughts and communicate information with other machines of the same kind.

8) Under the guidance of TCR, conscious machines should have some in-built reflexes play a role and employ the minimum requirement of perception to start with their bare-bones conscious experience of existence and non-existence about something. Then, the machines should get inspired from genetic epistemology to acquire physical knowledge gradually with many types of perception for more complex consciousness. Finally, they could use physical symbols to further develop and improve their conscious intelligence by inventing more abstract knowledge through actions, and by learning social knowledge from other machines of the same kind. Note that the physical forms of knowledge should be closely related to their perceiving sensors and responding effectors.

## 7. Conclusions

In this article, two first principles, namely, the Principle of World's Relativity and the Principle of Symbol's Relativity, have been proposed to elucidate the nature of intelligence comprehensively at the system level. Notably, the two first principles are fundamental as well as compatible in all phenomena of intelligence. They are independent of physics, chemistry and biology, and cannot be reduced to any principles of these sciences. In fact, they can be arguably regarded some "sciphi" perspectives, which belong to a theoretical level between science and philosophy. On the one hand, the two principles allow an intelligent agent to have many perceiving sensors implemented in different sciences, meaning that they are beyond science; on the other hand, the two principles can make not only speculative philosophical explanations, but also testable scientific predictions (e.g. radio language in the thought experiment "robonauts in space exploration"), meaning that they are below philosophy. Moreover, their importance and necessity have been shown by thought experiments to solve the mind-body problem. Hence, they have significances to establish a scientific theory of mind and consciousness, which is becoming a new paradigm for intelligence science. This is also a "sciphi" paradigm, called the Theory of Cognitive Relativity, with an abbreviation of TCR.

Rather than brain-like intelligence, the TCR indeed advocates a promising approach to true AI, especially with artificial consciousness different from humans' and animals'. Like theoretical physics (e.g. Newtonian mechanics, quantum mechanics and theory of relativity) with a core of first principles, the core of the TCR is defined as a simple and elegant set of first principles, at least including the Principle of World's Relativity and the Principle of Symbol's Relativity. The Principle of World's Relativity can bridge between the subjective world of an intelligent agent and the objective world it perceives through physical interactions with sensors and effectors. The Principle of Symbol's Relativity can help design physical forms of language for the agent to think and communicate appropriately. From the point of TCR's view, a computational model of brain will not be sufficient to explain the consciousness of human brain. The best explanation of a brain's consciousness is the brain itself, not its model. Even if a model of a brain has a consciousness, the model's consciousness may not be the brain's consciousness at all. In practice, true AI should center on realization of intelligent machines with subjective worlds that can be different from the human world and the animal worlds. These machines may have no vision, no hearing, no smell, no taste, and no touch. Instead, they can have people-different senses, such as of polarized light, electric field, magnetic field, ultrasound, infrasound, ultraviolet and infrared. Moreover, they can think and communicate in radio language rather than sound language. In short, true AI can be realized in a brain-different way, though it may be inspired by neuroscience (Hassabis et al. 2017). The brain-like intelligence tries to achieve intelligence as demonstrated by brains (Sendhoff et al. 2009), preferably of highly evolved creatures. However, the nature of intelligence may have something similar to the secret of flight. The flight of an airplane does not need to flap its wings like a bird. This bird-different flight is based on aerodynamics, not on imitation of the biological bird. Logically, aerodynamics can be described without biological terms. Therefore, intelligence science should be expected a theory independent of the biological brain. Without biological terms, the science would be able to explain how an agent's conscious intelligence arises from its physical interactions between the subjective world and the objective world. Furthermore, the science would make a guide to diverse realizations of true AI in a brain-different way. Finally, based on the intelligence science, the insights for true AI and the creed for consciousness studies, the TCR would probably drive an intelligence revolution in combination with some additional first principles.

## Acknowledgments

I am grateful to Hongwei Mo and Xiaochu Zhang for helpful comments on early versions of this article. This work was supported by the National Natural Science Foundation of China under grant 61876010.

# NOTES

1. The ampullae of Lorenzini are special sensing organs called electroreceptors, forming a network of jelly-filled pores. They are mostly discussed as being found in cartilaginous fish (sharks, rays, and chimaeras). For more information, please access https://en.wikipedia.org/wiki/Ampullae_of_Lorenzini.

2. Grapes are known to be highly toxic to dogs, though research has yet to pinpoint exactly which substance in the fruit causes this reaction. Please access https://www.akc.org/expert-advice/nutrition/can-dogs-eat-grapes/.

3. The principle of relativity is the requirement that the equations describing the laws of physics have the same form in all admissible frames of reference. Please access https://en.wikipedia.org/wiki/Principle_of_relativity.

4. Please access https://en.wikipedia.org/wiki/Brain_in_a_vat, for more information about the "brain in a vat", including Figure 5.

5. The self-refuting argument, "they are brains in a vat," or "we are brains in a vat", could be found in the article at https://philosophy.as.uky.edu/sites/default/files/Brains%20in%20a%20Vat%20-%20Hilary%20Putnam.pdf.

6. This definition of consciousness could be found at https://www.britannica.com/topic/consciousness.

7. Kirobo is the first talking robot astronaut in the world, tasked to be a companion. More information available at http://www.telegraph.co.uk/news/science/space/10221399/Talking-robot-astronaut-blasts-into-space.html.

8. Please access https://en.wikipedia.org/wiki/Computational_theory_of_mind, for more information about the computational theory of mind.

9. Functionalism of the mind claims that mental states (e.g. beliefs and desires) are constituted solely by their functional role. That is, they have causal relations to other mental states, numerous sensory inputs, and behavioral outputs. More information available at https://en.wikipedia.org/wiki/Functionalism_(philosophy_of_mind).